\documentclass[runningheads]{llncs}

\usepackage{eccv}

\usepackage{eccvabbrv}

\usepackage{graphicx}
\usepackage{booktabs}

\usepackage[accsupp]{axessibility}  % Improves PDF readability for those with disabilities.

\usepackage{hyperref}

\usepackage{orcidlink}

\usepackage{tabularray}
\usepackage{multirow}
\usepackage{wrapfig}

\usepackage{placeins}

\begin{document}

\title{SG-NeRF: Neural Surface Reconstruction with Scene Graph Optimization} 

\titlerunning{SG-NeRF}

\author{
Yiyang Chen$^{1,2*}$ \quad
Siyan Dong$^{3*}$ \quad
Xulong Wang$^{1}$ \quad
Lulu Cai$^{1,4}$ \\
Youyi Zheng$^{1\dagger}$ \quad
Yanchao Yang$^{3,4\dagger}$
}

\def \thefootnote{*}\footnotetext{Equal contributions (chen\_yy@zju.edu.cn, siyan3d@hku.hk).}
\def \thefootnote{$\dagger$}\footnotetext{Corresponding authors.}
\def \thefootnote{2}\footnotetext{This work was done during the author's internship at Chohotech Co. ltd..}

\authorrunning{Y.~Chen \& S.~Dong et al.}

\institute{
$^{1}$ State Key Lab of CAD\&CG, Zhejiang University $^{2}$ Chohotech Co. ltd.\\
$^{3}$ Institute of Data Science, The University of Hong Kong \\ 
$^{4}$ Department of Electrical and Electronic Engineering, The University of Hong Kong
}

\maketitle

\begin{abstract}
3D surface reconstruction from images is essential for numerous applications. 
Recently, Neural Radiance Fields (NeRFs) have emerged as a promising framework for 3D modeling. 
However, NeRFs require accurate camera poses as input, and existing methods struggle to handle significantly noisy pose estimates (i.e., outliers), which are commonly encountered in real-world scenarios. 
To tackle this challenge, we present a novel approach that optimizes radiance fields with scene graphs to mitigate the influence of outlier poses. 
Our method incorporates an adaptive inlier-outlier confidence estimation scheme based on scene graphs, emphasizing images of high compatibility with the neighborhood and consistency in the rendering quality. 
We also introduce an effective intersection-over-union (IoU) loss to optimize the camera pose and surface geometry, 
together with a coarse-to-fine strategy to facilitate the training. 
Furthermore, we propose a new dataset containing typical outlier poses for a detailed evaluation. 
Experimental results on various datasets consistently demonstrate the effectiveness and superiority of our method over existing approaches, showcasing its robustness in handling outliers and producing high-quality 3D reconstructions. 
Our code and data are available at: 
\url{https://github.com/Iris-cyy/SG-NeRF}. 

\end{abstract}

\begin{figure}[!t] 
\centering
\includegraphics[width=1.0\linewidth]{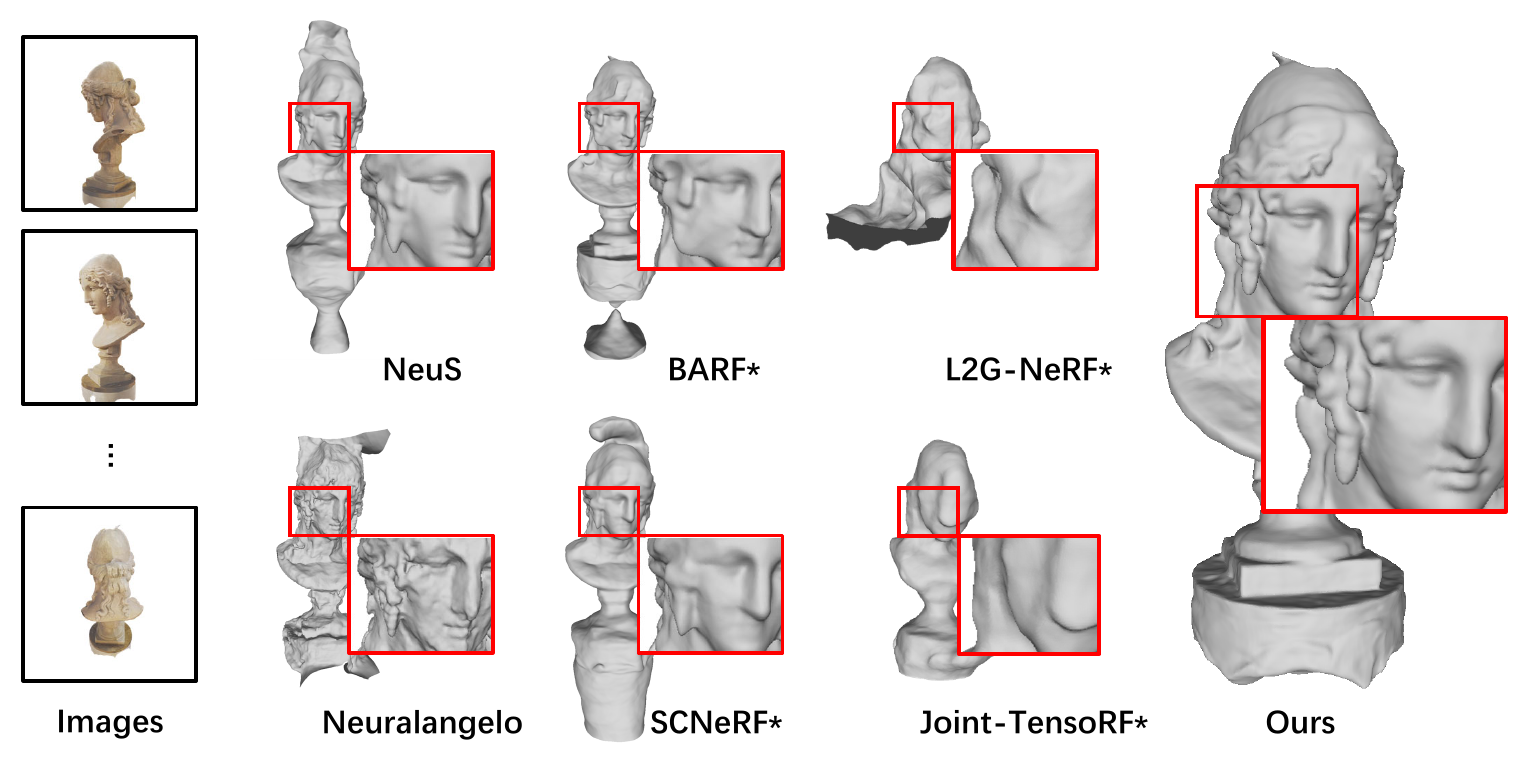}
\caption{
3D surface reconstruction (meshes) from images with camera poses that present significant noise. 
Directly training radiance fields with noisy poses can lead to incorrect structures (NeuS~\cite{wang2021neus} and Neuralangelo~\cite{li2023neuralangelo}). 
Recent approaches that focus on optimizing camera poses (BARF~\cite{lin2021barf}*, SCNeRF~\cite{jeong2021self}*, L2G-NeRF~\cite{chen2023local}*, and Joint-TensoRF~\cite{cheng2024improving}*, where * denotes their integration of NeuS for surface modeling) also fall short in handling significant pose errors, 
leading to unsatisfactory reconstruction. 
Our method works effectively and can produce high-quality 3D reconstructions.
}
\label{fig:teaser} 
\end{figure}

\section{Introduction}
\label{sec:intro}

3D mapping and reconstruction from multi-view images is crucial for a wide range of applications, such as virtual and augmented reality. 
Given a set of unorganized images captured around an object, most pipelines proceed in two stages for obtaining the reconstruction. 
Firstly, Structure-from-Motion (SfM) techniques~\cite{hartley2003multiple,schoenberger2016sfm} are employed to estimate camera poses of the images and produce a sparse scene representation. 
Consecutively, the 3D scene geometry can be recovered using Multi-View Stereo (MVS) algorithms~\cite{schoenberger2016mvs,furukawa2015multi,yao2018mvsnet,gu2020cascade,wang2022itermvs}, which assume accurate camera poses and derive the dense reconstruction.

Recently, Neural Radiance Fields (NeRFs)~\cite{mildenhall2020nerf} have been proposed for high-quality image synthesis.
The core idea of NeRFs is to implicitly encode a set of posed images into the weights of a neural network. 
The resulting implicit scene representations~\cite{barron2022mip,tancik2023nerfstudio} have demonstrated photo-realistic rendering quality and the capability for novel view synthesis (NVS). 
Given the connection between NVS and 3D reconstruction, 
NeRFs have been further adapted to develop surface-aware representations~\cite{wang2021neus,li2023neuralangelo}, enabling high-quality 3D reconstruction.

Despite the promising results, one tension of scene reconstruction with NeRFs is the dependence on accurate camera pose estimates. 
In practice, NeRF and many variants employ COLMAP~\cite{schoenberger2016sfm}, a widely used SfM framework, to estimate camera poses before training the scene representation. 
Unfortunately, the poses obtained could have significant errors affecting NeRF's reconstruction quality. 
Therefore, recent works~\cite{wang2021nerf,chng2022gaussian,bian2023nope,yan2023cf}
perform joint optimization of the scene representation and camera poses to mitigate the effect. 
However, these works still struggle with outlier poses, e.g., they either focus on forward-facing cameras with relatively small baselines or require proper initial poses for water-tight meshes in inward-facing scenarios~\cite{lin2021barf,jeong2021self,truong2023sparf,chen2023local,cheng2024improving}.

Given the context, an outlier refers to an image with a noticeably incorrect camera pose estimation, which can hardly be rectified via local optimization. 
Outlier images can happen when repetitive patterns or textureless regions are present, resulting in SfM failures. 
On the other hand, SfM systems provide byproducts such as sparse 3D reconstructions and scene graphs besides camera poses. 
Several works~\cite{kangle2021dsnerf,roessle2022dense} have shown that leveraging the sparse reconstruction can lead to faster training using fewer images. 
However, to the best of our knowledge, previous research has yet to explore using scene graphs to further optimize camera poses for better geometry reconstruction.

In this paper, we propose a novel framework that jointly optimizes the neural radiance field with a scene graph to alleviate the influence of outliers. 
We also initialize the scene graph with an SfM system~\cite{schoenberger2016sfm,detone2018superpoint,sarlin2020superglue}. 
To facilitate the joint training, we introduce an adaptive inlier-outlier confidence estimation mechanism to account for the influence of outlier images/poses.
Besides NeRF's photometric loss, 
we propose an intersection-over-union (IoU) loss across paired images in the scene graph, whose selection is coupled with the estimated confidence, to further optimize camera poses. 
Additionally, we apply a coarse-to-fine strategy to ensure the stability and efficiency of the training process. 
To perform a thorough evaluation, we collect a new dataset of 8 challenging scenes containing various levels of incorrect camera poses, as estimated by our SfM system.
We conduct experiments on both the proposed and existing datasets from the literature.
The experiments demonstrate the effectiveness of our method, highlighting its ability to produce high-quality 3D reconstruction results while remaining robust in the presence of outlier images or poses (e.g., Figure~\ref{fig:teaser}).
To summarize: 
\begin{itemize}

\item 
We investigate a practical problem of NeRF-based 3D reconstruction from images with significant pose errors. 
In contrast to previous works that assume moderate pose errors, we aim for a more challenging yet practical scenario. 
The images are casually captured without being carefully selected, which can lead to failures of state-of-the-art SfM systems.

\item 
Accordingly, we propose a novel method that performs a joint optimization of the radiance field and the scene graph initialized by an SfM. 
The proposed can reconstruct 3D surface under significant camera pose noise with an adaptive inlier-outlier confidence estimation, an IoU loss that efficiently leverages the confidence for pose correction, and a coarse-to-fine strategy that effectively promotes the training.

\item 
Besides showing better reconstruction performance on existing datasets, e.g., DTU~\cite{jensen2014large}, our method also achieves state-of-the-art results on a newly proposed dataset for multiview 3D that presents significant outlier camera poses. 

\end{itemize}

\section{Related Work}
\label{sec:related}

\paragraph{Neural radiance fields (NeRFs). }
The central concept of NeRFs~\cite{mildenhall2020nerf,barron2022mip,tancik2023nerfstudio,kaizhang2020nerfpp,mueller2022instant,xu2023jacobinerf,zhu2023vdn} 
is to represent scenes as implicit fields. 
Such representation is obtained by separately training for each scene with posed images. 
Integrating the signed distance function (SDF)~\cite{curless1996volumetric,park2019deepsdf,yariv2020idr} 
into NeRFs~\cite{wang2021neus,Oechsle2021unisurf,yariv2021volsdf,wang2022hf,fu2022geo, wu2022voxurf,wang2023neus2,cai2023neuda,li2023neuralangelo} 
allows the representation to learn implicit surfaces, thus enabling 3D mesh reconstruction. 
Due to simplicity and efficiency, we choose NeuS~\cite{wang2021neus} as our NeRF representation. 

Since NeRFs require known camera poses as input, 
SfM techniques are usually applied to register the images before NeRF training. 
Besides camera poses, SfMs produce scene graphs and sparse 3D reconstructions. 
Several works~\cite{kangle2021dsnerf,roessle2022dense} indicate that the sparse reconstructions can offer several enhancements compared to plain NeRFs. 
However, previous research has not yet investigated the usage of scene graphs for camera pose optimization. 

There are also works that perform joint NeRF and camera pose optimization with modular modifications~\cite{wang2021nerf,lin2021barf,chng2022gaussian,heo2023robust,park2023camp,cheng2024improving,bian2024porf} and cross-view correspondences~\cite{jeong2021self,truong2023sparf}. 
Most assume that all images have poses initialized properly and aim at local optimization for pose correction, e.g., L2G-NeRF~\cite{chen2023local} presents a local-to-global registration pipeline to alleviate the suboptimality. 
However, they still suffer from the presence of outlier images. 
In this paper, we leverage scene graphs and introduce an adaptive inlier-outlier confidence scoring mechanism to mitigate the influence of outlier images in scene reconstruction. 
In contrast to~\cite{jeong2021self,truong2023sparf}, 
we propose an intersection-over-union (IoU) loss, further enhancing the geometry quality of the implicit surface. 
Existing works also perform joint camera tracking and scene reconstruction in a sequential fashion~\cite{bian2023nope,zhu2022nice,yan2023cf}, which go beyond the scope of our work and we omit due to limited space.

\paragraph{Structure-from-Motion (SfM) and (re)localization.}
SfM techniques~\cite{hartley2003multiple} are widely used as data pre-processing for NeRF reconstruction. 
Given a set of unorganized images, SfM systems~\cite{snavely2008modeling,wu2011visualsfm,schoenberger2016sfm,theia-manual,lindenberger2021pixsfm,Liu_2023_LIMAP} organize the images by estimating camera poses and triangulating 3D scene points. 
Global SfM approaches~\cite{theia-manual,cui2015global} are time efficient, yet can be sensitive to outliers. 
Incremental SfM systems~\cite{snavely2006photo,schoenberger2016sfm} are more commonly used in the literature on NeRFs. 
Following previous research~\cite{lin2021barf,jeong2021self}, we utilize COLMAP~\cite{schoenberger2016sfm} as a basic component of our SfM module. 
It estimates camera poses along with a graph structure, i.e., scene graph, where each node represents an image and each edge indicates the presence of estimated matching points in the connected images. 
Despite the usage of geometric verification~\cite{torr1997assessment,frahm2006ransac} and outlier filtering~\cite{schoenberger2016sfm} in SfMs, it is highly possible for the scene graphs to still contain significantly incorrect pose estimations (i.e., outliers). 
The presence of outlier poses can heavily affect NeRF training, resulting in incorrect geometry and blurred appearance.

The task of localization~\cite{sattler2016efficient,sarlin2019coarse} is closely related to incremental SfM.
Given a set of posed images as a database, it aims to estimate the camera poses of new images captured from different viewpoints. 
Recent studies~\cite{moreau2022lens,moreau2023crossfire} have introduced NeRFs for the visual localization task. However, their pose accuracy still falls behind.
Although there are implicit representations~\cite{brachmann2021visual,dong2021robust} that can achieve better pose estimation, they can hardly be extended to optimize the poses within the database. 
In the proposed method, we leverage the recent advances of hloc~\cite{sarlin2019coarse,sarlin2020superglue}, an SfM-based visual localization toolbox, to supplement our SfM module with SuperPoint~\cite{detone2018superpoint} and SuperGlue~\cite{sarlin2020superglue}.

\section{Method}
\label{sec:method}

In the following, we first detail the problem setting and provide an overview of the proposed pipeline.
Then, we elaborate on the key technical designs in the subsequent sections. 

\paragraph{Problem statement.}
We aim for 3D surface reconstruction of object-level scenes from unorganized image sets. 
We assume that the camera's intrinsic parameters are known and that there are no distortions in images.
We focus on inward-facing scenes, a common scenario for object scanning in practice.
Specifically, for each scene, the input is a set of RGB images $\mathbf{I} =\{ I_1, I_2, ..., I_n \}$, and the output is a 3D surface reconstruction $S$ of the scene. 
The byproduct of our method is the optimized camera pose $P_i =(R_i, t_i)$ for each training image, where $R_i \in SO(3)$ and $t_i \in \mathbb{R}^3$ denoting the rotation and translation respectively. 
Each pose is assigned an inlier-outlier confidence score. One can also synthesize novel view images from the trained radiance field.

\begin{figure*}[!t] 
\centering
\includegraphics[width=1.0\linewidth]{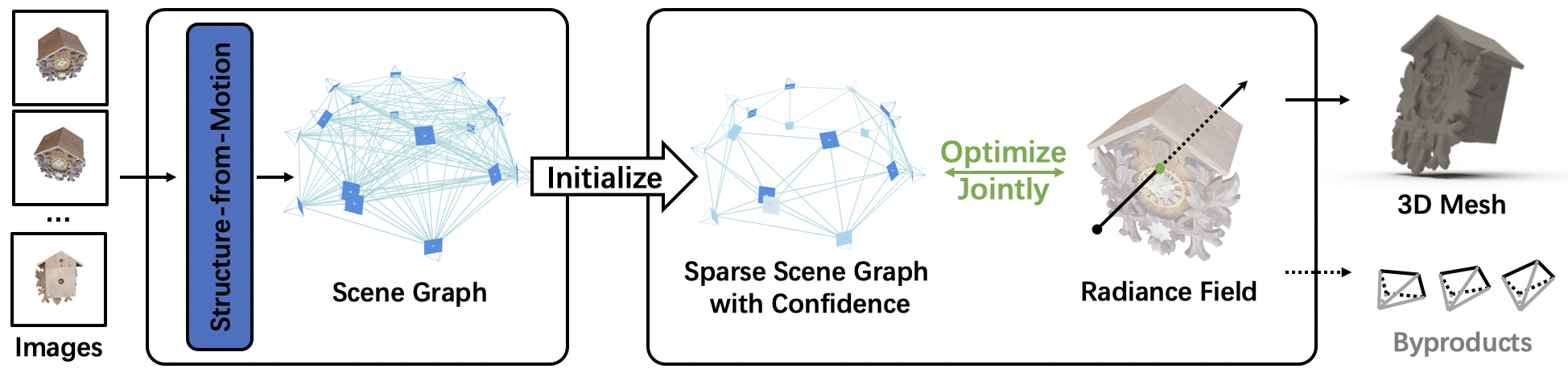}
\caption{
An overview of the proposed joint learning pipeline. 
Given a set of images, we first apply a Structure-from-Motion (SfM) algorithm to construct an initial scene graph (left), within which, each node represents a posed image. 
An edge between two nodes suggests that the involved images are to share overlapped regions. 
Next, the initial scene graph is sanctified. 
Each node is then assigned a confidence score based on the number of matching points among neighboring nodes. 
Then, we train a Neural Radiance Field (NeRF) using the confidence-aware scene graph and images. 
The training process alternates between fitting the radiance field and updating the scene graph. 
Eventually, we can extract the 3D scene mesh from the trained field. 
}
\label{fig:pipeline} 
\end{figure*}

\paragraph{Method overview.}
Figure~\ref{fig:pipeline} illustrates the workflow of the proposed pipeline. 
Given the training images, we first apply a widely used Structure-from-Motion (SfM) algorithm, i.e., COLMAP~\cite{schoenberger2016sfm}, to construct an initial scene graph of the images,
where the keypoint descriptor and matching are provided by SuperPoint~\cite{detone2018superpoint} and SuperGlue~\cite{sarlin2020superglue}, respectively. 
The scene graph created through SfM usually contains outlier poses. 
Therefore, we refine the graph and allocate an inlier-outlier confidence score to each node. 
Following this, we train a Neural Radiance Field (NeRF) using the refined scene graph. 
The training process is essentially a scene-specific joint optimization. 
It involves alternating between adjusting the radiance field and updating the scene graph. 
In particular, the radiance field learns to recover the 3D densities and RGB colors in the scene. Simultaneously, the scene graph optimizes camera poses and their confidence scores, gradually eliminating the influence of estimated outliers.
After training, we extract the 3D scene mesh from the density of the optimized radiance field.

Next, we describe how to initialize the scene graph (Sec.~\ref{subsec:posegraph}). 
Then, we present our joint optimization method for training the radiance field and updating the scene graph (Sec.~\ref{subsec:loss}). 
Lastly, we introduce a coarse-to-fine training strategy to ensure an efficient and stable training process (Sec.~\ref{subsec:coarse2fine}).

\subsection{Scene Graph}
\label{subsec:posegraph}
A scene graph $G = (V,E)$ in SfM consists of a set of nodes $V$ and edges $E$. 
Each node $v_i \in V$ corresponds to an input image $I_i\in\mathbf{I}$, and an edge between two nodes indicates that the connected images share a co-visible region of the scene. 
More explicitly, 
the nodes record the camera poses $\{ P_1, P_2, ..., P_n \}$ of the corresponding images, 
and the edges record the keypoint matches 
$\mathbf{M} = \{ M_{i,j} | I_i, I_j \in \mathbf{I} \}$ of the paired images $I_i, I_j$, i.e., 
$M_{ij}= \{ (kp_i^{(n)}, kp_j^{(n)})| kp_i^{(n)}, kp_j^{(n)} \in \mathbb{R}^2 \}$ 
is the set of matched keypoint locations $kp_i^{(n)}, kp_j^{(n)}$ ($n$-th match between $I_i$ and $I_j$).

\paragraph{Initialization.} 
The scene graph is initially constructed with the employed SfM module~\cite{schoenberger2016sfm}, which contains two major steps: a) correspondence search, and b) incremental registration and reconstruction. 
In the first step, 
we apply pre-trained SuperPoint~\cite{detone2018superpoint} to extract keypoints from images, and exhaustively match every pair of images with keypoints using pre-trained SuperGlue~\cite{sarlin2020superglue} model. 
As a result, we obtain a set of successfully matched image pairs. 
Since there could be incorrect matches, the SfM pipeline validates each pair by estimating a relative pose via solving the essential matrix~\cite{hartley2003multiple} with RANSAC~\cite{fischler1981random} from a set of putative matches. 
If a valid essential matrix exists that maps a sufficient number of keypoints between the two images, the pair is considered to have passed the verification. 
The initial scene graph is constructed by setting the images as nodes and the verified pairs as edges.

The second step begins with selecting an appropriate image pair to initialize a metric reconstruction as well as the global coordinate system of the scene, which triangulates 3D coordinates from keypoints. 
Subsequently, the SfM system alternates between image registration and scene reconstruction. 
An image $I_i$ is registered by estimating an absolute camera pose $P_i$ in the scene coordinate system, which is then recorded in the corresponding node $v_i$. This is achieved by solving the Perspective-n-Point (PnP) problem~\cite{lepetit2009ep} using RANSAC~\cite{fischler1981random}.
Once an image is registered, it expands the 3D reconstruction by triangulating new keypoints. 
Following registration and triangulation, the SfM system performs bundle adjustment and employs a filter to detect and remove outlier keypoints.
The matches after the filtering process are recorded to the edges $E$.

\begin{figure}[!t] 
\centering
\includegraphics[width=0.98\linewidth]{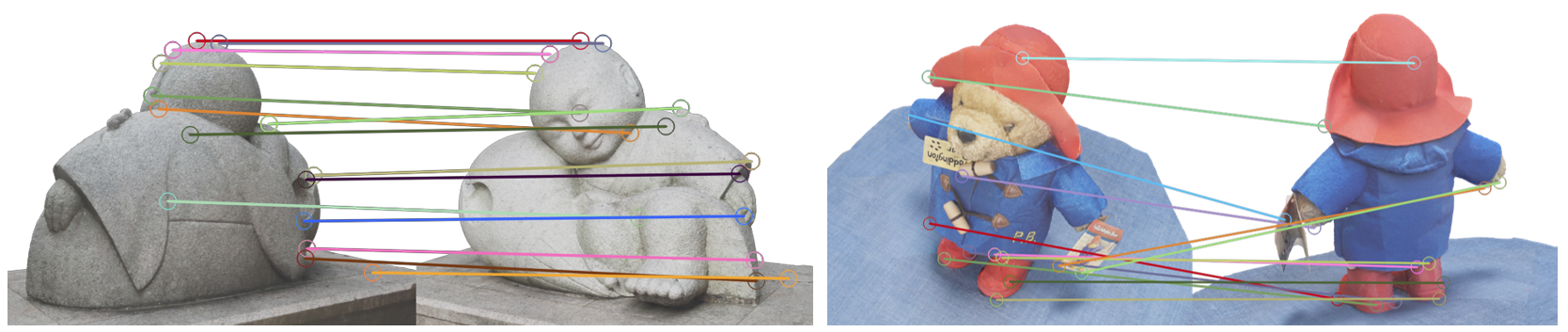}
\caption{
Visualization of matches that are falsely established as correspondences from non-overlapping regions. The results are obtained using COLMAP~\cite{schoenberger2016sfm} with SuperPoint~\cite{detone2018superpoint} and SuperGlue~\cite{sarlin2020superglue}. 
Note that most of the estimations are incorrect and can heavily affects the reconstruction quality. 
}
\label{fig:sfm_fail} 
\end{figure}

\paragraph{Pruning and confidence estimation.}
Despite the outlier filtering in the SfM process, the scene graph obtained may still contain errors that affect the reconstruction quality. 
As shown in Figure~\ref{fig:sfm_fail}, there can be false positive edges resulting from incorrect matches.
Empirically, a pair of images is less reliable when there is a larger relative rotation or fewer matches.
Therefore, we set an angular threshold $\tau$ for the estimated relative rotations and remove any edges exceeding $\tau$. 
This sparsification effectively prunes low-quality image pairs (i.e., edges) and reduces redundancy. 
Furthermore, we assign a confidence estimate to each node based on keypoint matches, which helps us determine whether it is likely an inlier or outlier. 
The confidence score for a node $v_i$ is computed as:
\begin{equation}
CS(v_i) = \frac{\sum_{M_{i,j} \in \mathbf{M}_i} |M_{i,j}|}{|\mathbf{M}_i|}  , 
\end{equation}
where $\mathbf{M}_i \subset \mathbf{M}$ denotes the set of edges connected to $v_i$ and $|\cdot|$ is the number of elements in a set. 
Then, the confidence score is normalized among all the nodes via $CS(v_i) = CS(v_i) / \sum_{v_i \in V} CS(v_i)$. 
As such, a higher score indicates that the image has more keypoint matches and a higher probability of being an inlier.
The confidence scores form a probability distribution and it is used to guide the sampling of training data for the consecutive radiance field training process.

The sparsified scene graph, along with its confidence estimates, are jointly optimized during the upcoming radiance field training process. 
Explicitly, the confidence scores will be updated, the camera poses will be optimized, and the graph structure will remain fixed.
Details are elaborated on in the next section.

\subsection{Joint Optimization} 
\label{subsec:loss}

Once we obtain a scene graph with confidence scores, we train a neural radiance field to fit the 3D scene representation. 
To perform 3D reconstruction, we utilize a neural implicit surface representation, NeuS~\cite{wang2021neus}, as our backbone. 
Below, we first briefly review the radiance field representation and then introduce our joint optimization scheme.

\paragraph{Radiance field.}
The neural radiance fields (NeRFs) ~\cite{mildenhall2020nerf} represent a scene by modeling the occupancy and color of a point in the 3D space. This is achieved by training a neural network to fit each individual scene. 
The network takes a 3D location and viewing direction as input and generates the corresponding density and RGB color (i.e., radiance) as output. The volume rendering technique allows the synthesis of an image by integrating radiance along the viewing ray. 
During the training process, the goal is to minimize the difference between synthesized pixels and those in real images as an L1 photometric loss:
\begin{equation}
\label{eq:photo}
\mathcal{L}_{photo} = |\hat{I}_i - I_i| .
\end{equation}

In order to extract high-quality surfaces from the field, NeuS~\cite{wang2021neus} enhances the density estimation with a signed distance function $f$ (SDF) and introduces an additional regularization loss on each viewing ray: 
\begin{equation}
\label{eq:reg}
\mathcal{L}_{reg} = \frac{1}{k} \sum_{i=1}^k (||\nabla f(p_i)||_2 -1)^2 , 
\end{equation}
where $f(p_i)$ represents the distance estimate for each sampled 3D location along the ray. 
We utilize NeuS as the backbone for our radiance field and leverage the two aforementioned loss terms during our training process.

\paragraph{Joint optimization of the field and scene graph.}
Our training process simultaneously fits the radiance field parameters and optimizes the scene graph. 
It consists of several training epochs.
In each epoch, we alternate between two steps: optimizing the parameters of the radiance field and camera poses ({\it field-pose step}), and updating the confidence scores ({\it confidence step}).

\begin{figure}[!t] 
\centering
\includegraphics[width=1.0\linewidth]{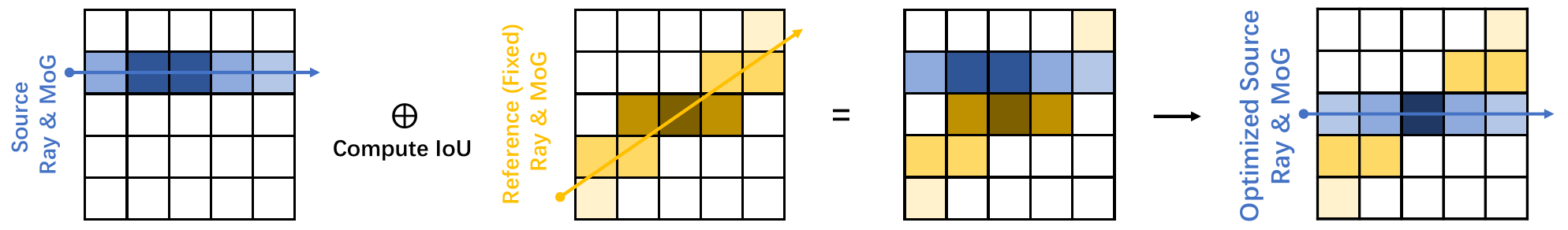}
\caption{
Illustration of the two-view intersection-over-union (IoU) loss in 2D that can be easily extended into 3D. 
Given a pair of matched keypoints from source and reference images, in order to maximize the IoU between the two rays, both the camera pose of the source image and the estimated density in the radiance field have to be optimized.  
}
\label{fig:iou} 
\end{figure}

During the {\it field-pose step,} 
we start by creating a temporary training set that includes images and their camera pose estimates. 
This set is established by {\it sampling with replacement based on the confidence scores}, which effectively loops in the adaptive confidence estimation into the NeRF training process. 
A higher confidence score indicates a larger probability of the corresponding image being selected. 
After selection, for the training images, besides photometric (Eq.~\ref{eq:photo}) and regularization (Eq.~\ref{eq:reg}) loss terms, we propose an intersection-over-union (IoU) loss.
As illustrated in Figure~\ref{fig:iou}, 
it is computed on top of keypoint matches. 
For each keypoint in a match, we project a ray from the camera center. 
We then fit a mixture of Gaussians (MoG) using the points sampled along this ray. The IoU loss aims to maximize the intersection-over-union between the two MoGs that correspond to the matched keypoints. It is calculated as: 
\begin{equation}
\mathcal{L}_{iou} = 1 - \frac{MoG(kp_i) \cdot MoG(kp_j)}{MoG(kp_i) + MoG(kp_j)}  .
\end{equation}
Given a training image as a source, we traverse through all the paired reference images to compute the IoU loss. 
Finally, our total training loss is defined as:
\begin{equation}
\label{eq:loss}
\mathcal{L} = \mathcal{L}_{photo} + \alpha \mathcal{L}_{reg} + \beta \mathcal{L}_{iou}\,. 
\end{equation}
The photometric loss helps optimize the radiance field's color, geometry, and the camera poses, while the regularization terms further bias the training towards more robust and geometrically meaningful reconstructions.

In the {\it confidence step,} 
the confidence scores are updated adaptively according to the actual image reconstruction quality. 
Since the initial scores are derived from keypoint matches, they might lack a comprehensive understanding of the information contained in an image. 
Especially, we observe that the synthesized image from an outlier viewpoint tends to exhibit significant artifacts, due to bad alignment with the majority of the properly posed images. 
Therefore, we calculate the peak signal-to-noise ratio (PSNR) for each image in the scene graph against its rendering from the radiance field.
The PSNR values are normalized among all the nodes, serving as a fidelity of the matching, and fused into the confidence scores as
\begin{equation}
\label{eq:psnr}
CS(v_i) = CS(v_i) + \lambda PSNR(v_i). 
\end{equation}
Finally, we re-normalize the scores to maintain a probability distribution, thereby guiding the sampling process in the next epoch.
As training progresses, the confidence scores are gradually scaled before the aforementioned re-normalization step to promote convergence.

\subsection{Coarse-to-Fine Strategy}
\label{subsec:coarse2fine}

During training, we further apply a coarse-to-fine strategy to improve robustness and efficiency. 
Firstly, we remove high-frequency details from the input images by a smoothing, 
which allows for a quick fitting of a coarse radiance field representation of the scene. 
As discussed in the recent study~\cite{lin2021barf}, removing high-frequency bands also helps to avoid getting stuck in local minimas during training. 
Moreover, it facilitates the proposed PSNR-based confidence update, because low PSNR can be observed even in accurately posed images containing a large amount of high-frequency details. 
Thus, removing high-frequency details of images during the initial training epochs tends to put more optimization capacity on poses. 
As the training progresses, we gradually recover the high-frequency details.
More specifically,
the coarse-to-fine strategy is implemented by applying a Gaussian filter to the original input images at the beginning of each epoch. 
We initially set a large standard deviation $\sigma$ for the Gaussian kernel (coarsest scale), which gradually decreases as the training proceeds. 
When $\sigma<1$ (pixel), we stop the Gaussian filtering and use the original images as input (finest scale). 
Next, we validate the proposed pipeline and designs.

\section{Experiments}
\label{sec:exps}

We evaluate the effectiveness of our method through extensive experiments on various datasets, 
which includes a new inward-facing dataset containing significant outlier camera poses produced by the SfM system (Sec.~\ref{subsec:dataset}). 
Next, we describe the implementation details of the proposed (Sec.~\ref{subsec:impl}). 
We then report the comparisons with state-of-the-art methods on both the proposed dataset and a widely used benchmark, DTU dataset~\cite{jensen2014large} (Sec.~\ref{subsec:comp}).
Furthermore, we perform a series of ablation studies and analyses to verify the effectiveness of each proposed component (Sec.~\ref{subsec:abla}).

\subsection{Proposed Dataset}
\label{subsec:dataset}

To demonstrate the generality of the stated problem setting and test the efficiency of the proposed method, 
we collect 3D meshes from BlendedMVS~\cite{yao2020blendedmvs} and construct a new inward-facing dataset by uniformly sampling camera viewpoints in the hemisphere around each mesh. 
We select 8 representative scenes, and each of the selected scenes contain 18-45 training images, except the scene \textit{Clock}, which comprises 108 training images. 

We calculate the initial camera poses for these training views with our SfM module. 
The initial reconstruction result of each scene contains significant incorrect poses, with a proportion ranging from 1/9 to 1/3. 
Most of these poses tend to come with a large angular deviation and cannot be rectified through local optimization.
Due to limited space, please refer to the supplementary material for a detailed description and statistics of the dataset.

\subsection{Implementation Details}
\label{subsec:impl}

We employ COLMAP~\cite{schoenberger2016sfm} as our SfM framework. Following hloc~\cite{sarlin2019coarse}, we replace the keypoints and the matching module with SuperPoint~\cite{detone2018superpoint} and SuperGlue~\cite{sarlin2020superglue}, respectively.
We employ the off-the-shelf weights from their official repository. 
After SfM, to prune the scene graph, we set the angular threshold to $\tau=70$ degrees for our dataset. 
For the DTU dataset, we set $\tau=45$ degrees because the viewpoints are more densely sampled. 
We set $\lambda$ (Eq.~\ref{eq:psnr}) as 1.0 to balance the initial and updated confidence scores. 

We implement our radiance field based on NeuS~\cite{wang2021neus}. 
We follow the hierarchical sampling strategy in NeuS and set the batch size to 512, among which, we select 16 matched keypoints and use them to calculate the IoU loss. 
The MoG of each ray is calculated using 8 points with the highest densities. Specifically, we assign a 3D Gaussian to each point with the point coordinates as the mean and a fixed value of 0.1 as the covariance. Then, we fuse the Gaussians along the ray to form a MoG by a normalized weighted sum, where the weights come from the field densities estimated by NeuS. 
We discretize the MoG with a resolution of $64 \times 64 \times 64$.
We set the loss weights (Eq.~\ref{eq:loss}) to $\alpha=0.1$ and $\beta=0.2$.

We initialize the coarse-to-fine parameter $\sigma$ as $max(H, W) \times 0.02$, where $H$ and $W$ represent the height and weight of the input images. 
After training, we use the marching cube algorithm~\cite{lorensen1998marching} to extract a 3D mesh from the radiance field.
All of the experiments are conducted on NVIDIA RTX 3090 GPUs.
Our method runs in average 11 hours for 150k iterations on the proposed dataset, and 18 hours for 300k iterations on the DTU dataset.

\subsection{Comparisons}
\label{subsec:comp}

We compare our method with existing approaches on the proposed dataset and the DTU dataset. 
We follow the evaluation protocol in the literature~\cite{wang2021neus,li2023neuralangelo} and report Chamfer distance and F-score for evaluating the mesh quality.

\begin{table*}[!htb]
\caption{
Quantitative results on our dataset.
The \textbf{\textcolor{red}{red}} and \textbf{\textcolor{blue}{blue}} numbers indicate the first and second performer for each scene. 
$\dagger$ denotes that only valid values are used for the average.
Overall, our method achieves the best reconstruction results. 
}

\centering
\resizebox{0.96\textwidth}{!}{ 
\begin{tabular}{cl|cccccccc|c}

\toprule
 & & Baby & Bear & Bell & Clock & Deaf & Farmer & Pavilion & Sculpture & Mean \\ 

\midrule
 
\multirow{8}{*}{\rotatebox{90}{Chamfer distance $\downarrow$}}  
& NeuS~\cite{wang2021neus} & 0.69 & 0.31 & 3.33 & 1.16 & 0.55 & 2.49 & 0.29 & 0.66 & 1.18 \\ 
& Neuralangelo~\cite{li2023neuralangelo} & 0.70 & 0.65 & - & 0.38 & 0.59 & 4.89 & 1.95 & \textbf{\textcolor{blue}{0.31}} & 1.35$^\dagger$ \\ 
& BARF~\cite{lin2021barf}* & 1.08 & 0.28 & 3.31 & \textbf{\textcolor{blue}{0.19}} & 0.46 & 2.13 & 0.38 & 0.57 & \textbf{\textcolor{blue}{1.05}} \\ 
& SCNeRF~\cite{jeong2021self}* & 1.19 & \textbf{\textcolor{blue}{0.27}} & 3.74 & 1.33 & 0.46 & \textbf{\textcolor{blue}{1.45}} & \textbf{\textcolor{blue}{0.23}} & 0.81 & 1.19 \\ 

& GARF~\cite{chng2022gaussian}* & 2.04 & 2.25 & 3.08 & 2.01 & 0.59 & 1.58 & 0.96 & 0.57 & 1.64 \\ 

& L2G-NeRF~\cite{chen2023local}* & 1.15 & 0.29 & \textbf{\textcolor{blue}{1.26}} & 0.24 & \textbf{\textcolor{blue}{0.40}} & 2.18 & - & 4.36 & 1.41$^\dagger$ \\

& Joint-TensoRF~\cite{cheng2024improving}* & 3.11 & - & 2.49 & 0.36 & 0.88 & 2.51 & 1.35 & 0.70 & 1.63$^\dagger$\\

& PoRF~\cite{bian2024porf} & \textbf{\textcolor{red}{0.31}} & 0.49 & - & - & \textbf{\textcolor{red}{0.30}} & 3.80 & 2.20 & - & 1.42$^\dagger$ \\

& SG-NeRF (Ours) & \textbf{\textcolor{blue}{0.56}} & \textbf{\textcolor{red}{0.25}} & \textbf{\textcolor{red}{0.98}} & \textbf{\textcolor{red}{0.15}} & 0.45 & \textbf{\textcolor{red}{0.87}} & \textbf{\textcolor{red}{0.20}} & \textbf{\textcolor{red}{0.22}} & \textbf{\textcolor{red}{0.46}} \\ 

\midrule

\multirow{8}{*}{\rotatebox{90}{F-score $\uparrow$}} 
& NeuS~\cite{wang2021neus} & 0.65 & \textbf{\textcolor{red}{0.93}} & 0.48 & 0.72 & 0.84 & 0.54 & 0.93 & 0.70 & 0.74 \\ 
& Neuralangelo~\cite{li2023neuralangelo} & 0.57 & 0.80 & - & 0.85 & 0.66 & 0.14 & 0.47 & \textbf{\textcolor{blue}{0.89}} & 0.63$^\dagger$ \\ 
& BARF~\cite{lin2021barf}* & 0.58 & 0.91 & 0.49 & \textbf{\textcolor{blue}{0.95}} & 0.86 & 0.51 & 0.86 & 0.87 & \textbf{\textcolor{blue}{0.75}} \\ 
& SCNeRF~\cite{jeong2021self}* & 0.56 & \textbf{\textcolor{red}{0.93}} & 0.49 & 0.69 & 0.86 & \textbf{\textcolor{blue}{0.59}} & \textbf{\textcolor{red}{0.95}} & 0.73 & 0.72 \\ 

& GARF~\cite{chng2022gaussian}* & 0.18 & 0.21 & 0.50 & 0.27 & 0.78 & 0.57 & 0.41 & 0.83 & 0.47 \\ 

& L2G-NeRF~\cite{chen2023local}* & 0.58 & \textbf{\textcolor{blue}{0.92}} & \textbf{\textcolor{blue}{0.65}} & 0.92 & \textbf{\textcolor{blue}{0.89}} & 0.49 & - & 0.21 & 0.67$^\dagger$ \\

& Joint-TensoRF~\cite{cheng2024improving}* & 0.20 & - & 0.38 & 0.84 & 0.60 & 0.24 & 0.34 & 0.63 & 0.46$^\dagger$ \\

& PoRF~\cite{bian2024porf} & \textbf{\textcolor{red}{0.92}} & 0.78 & - & - & \textbf{\textcolor{red}{0.92}} & 0.39 & 0.35 & - & 0.67$^\dagger$ \\

& SG-NeRF (Ours) & \textbf{\textcolor{blue}{0.74}} & \textbf{\textcolor{red}{0.93}} & \textbf{\textcolor{red}{0.71}} & \textbf{\textcolor{red}{0.96}} & 0.87 & \textbf{\textcolor{red}{0.76}} & \textbf{\textcolor{blue}{0.94}} & \textbf{\textcolor{red}{0.92}} & \textbf{\textcolor{red}{0.85}} \\ 

\bottomrule

\end{tabular}

}

\label{tab:ours}
\end{table*}

\begin{figure}[!htb] 
\centering
\includegraphics[width=0.98\linewidth]{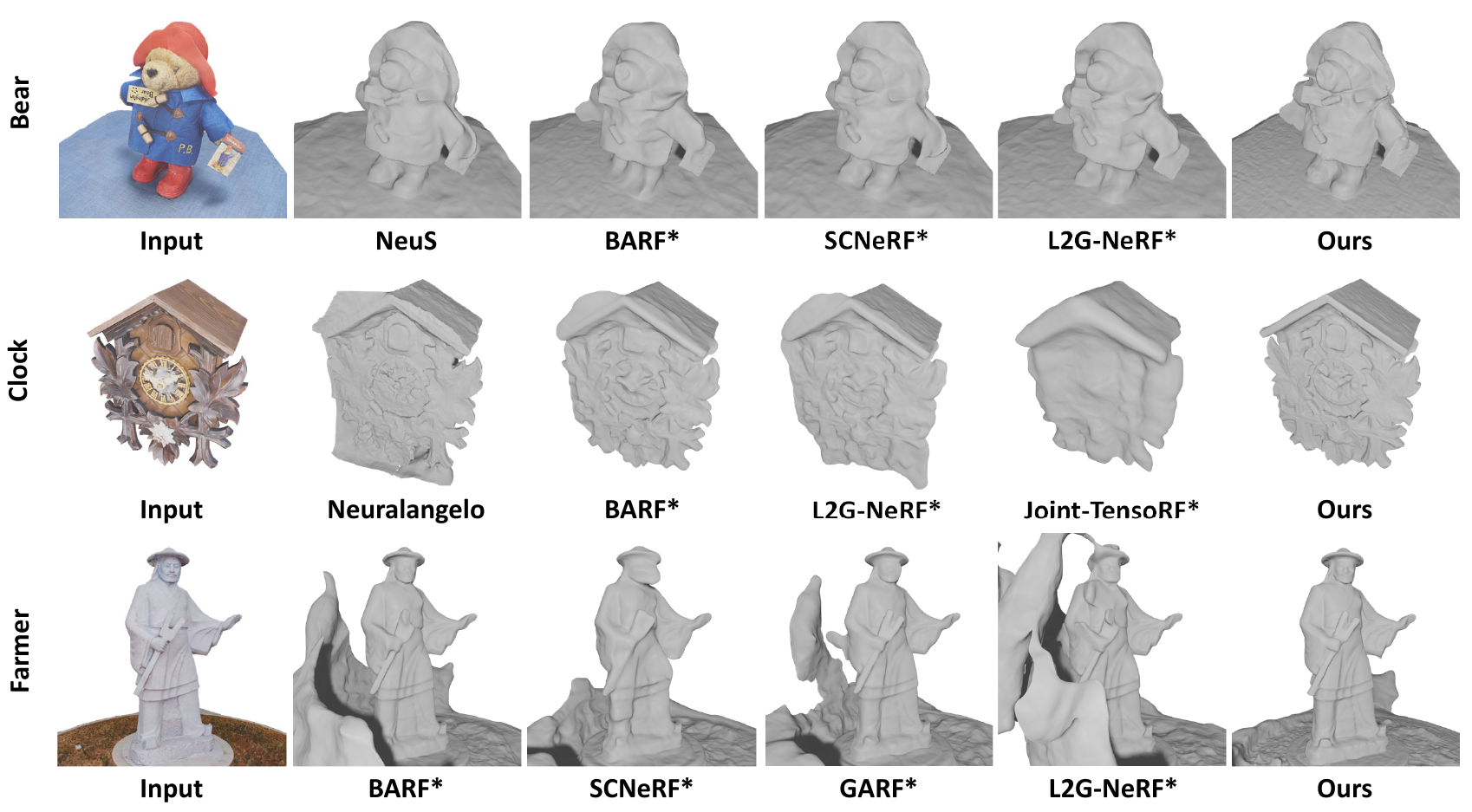}
\caption{
Qualitative comparisons on the proposed dataset.
As shown, our method is more robust to outlier poses,
producing less distortion and better geometric detail. 
For the sake of space, we display the five top-performing results for each scene.
}
\label{fig:ours} 
\end{figure}

\begin{figure}[!ht] 
\centering
\includegraphics[width=1.0\linewidth]{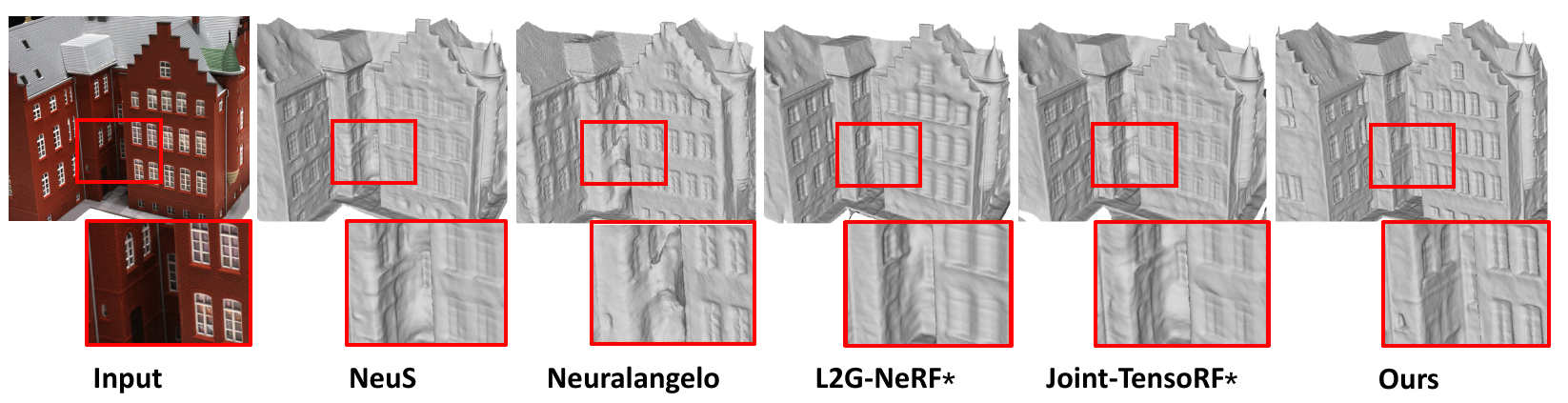}
\caption{
Qualitative comparison on the DTU dataset (top five). 
Neuralangelo shows detailed windows but struggles with the gap area between buildings. 
In contrast, our method produces more accurate details.
}
\label{fig:dtu} 
\end{figure}

\paragraph{Competitors.} 
We compare our method with our backbone model NeuS~\cite{wang2021neus} and the state-of-the-art Neuralangelo~\cite{li2023neuralangelo}. 
We integrate our backbone with state-of-the-art camera pose optimization methods to create a series of primary competitors. 
Specifically, we select BARF~\cite{lin2021barf}, SCNeRF~\cite{jeong2021self}, GARF~\cite{chng2022gaussian}, L2G-NeRF~\cite{chen2023local}, and Joint-TensoRF~\cite{cheng2024improving} as the baselines. 
For each of them, we adopt the official implementation to optimize camera poses, and then apply the optimized poses to train NeuS. The resulting competitors are respectively labeled as BARF*, SCNeRF*, GARF*, L2G-NeRF*, and Joint-TensoRF*.
Note that for all methods, we utilize the initial camera poses derived from our SfM module for a fair comparison. 
We also compare with recent work PoRF~\cite{bian2024porf} on our dataset. It jointly optimizes camera poses and neural surface reconstruction. 

\paragraph{Results on our dataset.}
The quantitative results are reported in Table~\ref{tab:ours}. 
Regarding NeuS and Neuralangelo, 
due to significantly noisy camera poses, 
their 3D reconstructions present significant flaws, as shown in Figure~\ref{fig:ours}. 
In contrast, our method shows robustness to pose errors and outperforms NeuS by 61\% in Chamfer distance and by 15\% in F-score.

Interestingly, among the competitors, the early one, BARF*, delivers the best overall performance. But it still lags behind our method. 
The subpar performance of the competitors is due to their pose optimization processes, namely, local optimizations, which cannot rectify the poses with significant errors. These incorrect poses can introduce large deviations during the optimization process, affecting the correct poses, and leading to an overall decline in the pose accuracy. 
On the contrary, thanks to the confidence-based sampling scheme, our method is more robust to outlier poses. 
Please also note that there are several failure cases from the competitors indicating completely incorrect reconstruction. For example, Neuralangelo in {\it Bell} and L2G-NeRF in {\it Pavilion}, while our method consistently achieves reasonable results across all scenes.

\begin{wraptable}{r}{0.5\textwidth}

\caption{
Quantitative results on the DTU dataset with noisy camera poses as input.
}
\centering

\resizebox{0.5\textwidth}{!}{
\begin{tabular}{l|ccccc|c} 
\toprule
Chamfer distance $\downarrow$ & 24 & 37 & 40 & 55 & 63 & Mean \\ 
\midrule

NeuS~\cite{wang2021neus} 
& 1.07 & 2.80 & 1.52 & 1.30 & 3.20 & 1.98 \\

Neuralangelo~\cite{li2023neuralangelo} 
& 1.06 & 2.96 & \textbf{\textcolor{blue}{1.22}} & \textbf{\textcolor{blue}{0.42}} & \textbf{\textcolor{blue}{1.23}} & \textbf{\textcolor{blue}{1.38}} \\

BARF~\cite{lin2021barf}*
& 1.46 & \textbf{\textcolor{red}{1.40}} & 5.16 & 1.78 & 1.80 & 2.32 \\

SCNeRF~\cite{jeong2021self}* 
& 1.45 & 2.84 & 2.60 & 0.78 & 1.83 & 1.90 \\

GARF~\cite{chng2022gaussian}*
& 1.18 & 2.00 & 2.61 & 2.37 & 8.74 & 3.38 \\

L2G-NeRF~\cite{chen2023local}* & 1.08 & \textbf{\textcolor{blue}{1.60}} & 3.27 & 1.79 & 6.97 & 2.94 \\

Joint-TensoRF~\cite{cheng2024improving}* & \textbf{\textcolor{blue}{1.00}} & 2.60 & - & - & 7.71 & 3.77$^\dagger$ \\

SG-NeRF (Ours) 
& \textbf{\textcolor{red}{0.87}} & 2.39 & \textbf{\textcolor{red}{0.88}} & \textbf{\textcolor{red}{0.38}} & \textbf{\textcolor{red}{1.13}} & \textbf{\textcolor{red}{1.13}} \\

\bottomrule
\end{tabular}
}

\label{tab:dtu}
\end{wraptable}

\paragraph{Results on the DTU dataset.}
The quantitative results are shown in Table~\ref{tab:dtu}. 
To simulate outlier poses, we randomly select 1/7 to 1/4 of the images for each scene, and inject random noise ($\epsilon_t \in [0, 90]$ degrees to the direction of the translation vector, and $\epsilon_r \in [0, 20]$ degrees to the rotation matrix) to the corresponding poses.
Since the ground-truth 3D models in DTU are obtained from structured light scanning, they have fine-grained details. 

As observed, Neuralangelo produces the second-best results. 
While BARF* achieves the best results in scene 37, it is more likely to impose negative impact on camera poses, thereby has worse performance in most scenes.
Compared to the competitors, our method achieves the best overall performance among all the methods. 
A qualitative comparison can be found in Figure~\ref{fig:dtu}.

\subsection{Analyses}
\label{subsec:abla}

\paragraph{Ablation studies.}
We select three representative scenes from the proposed dataset and conduct ablation studies to evaluate the effectiveness of each component. Specifically, the experiments aim to validate the contribution of the scene graph (joint optimization), the IoU loss, and the coarse-to-fine strategy.

The results are reported in Table~\ref{tab:ablation}.
To evaluate the effectiveness of the joint optimization, 
we directly train our method using the original scene graph obtained from SfM without further refinement. 
A noticeable performance drop is observed when compared to our full method (as indicated by w/o $\tau$ in the table). By disabling the use of the confidence score (w/o $CS$), 
an even larger performance drop is observed. 
Similar performance drop is also observed when we remove the IoU term from the loss function (w/o IoU). 
These highlight the importance of employing the proposed components in the joint optimization framework.
Finally, the usefulness of the coarse-to-fine strategy is verified by comparing with the second last column (w/o C2F).

\begin{table*}[tbh!]
\caption{
Quantitative results of our ablation studies. 
We individually remove the use of sparsification by thresholding (w/o $\tau$), confidence estimation (w/o $CS$), Intersection-over-Union loss (w/o IoU), and coarse-to-fine optimization strategy (w/o C2F) from our full method.
The \textbf{bold} numbers indicate the best results. }

\centering
\resizebox{0.96\textwidth}{!}{

\begin{tabular}{l|ccccc|ccccc}

\toprule

& \multicolumn{5}{c|}{Chamfer distance $\downarrow$}        
& \multicolumn{5}{c}{F-score $\uparrow$} \\

& w/o $\tau$ & w/o $CS$ & w/o IoU & w/o C2F & Full
& w/o $\tau$ & w/o $CS$ & w/o IoU & w/o C2F & Full \\

\midrule

Bell      & 1.27 & 1.32 & 1.68 & 1.15 & \textbf{0.98} 
& 0.56 & 0.64 & 0.40 & 0.67 & \textbf{0.71} \\

Pavilion  & 0.27 & 0.30 & 0.26 & 0.21 & \textbf{0.20} 
& 0.90 & 0.88 & 0.91 & 0.93 & \textbf{0.94} \\

Sculpture & 0.26 & 0.38 & 0.23 & 0.37 & \textbf{0.22} 
& 0.91 & 0.89 & 0.91 & 0.85 & \textbf{0.92} \\

\midrule

Mean & 0.60 & 0.67 & 0.72 & 0.58 & \textbf{0.47} 
& 0.79 & 0.80 & 0.74 & 0.82 & \textbf{0.86} \\ 

\bottomrule

\end{tabular}
}

\label{tab:ablation}
\end{table*}

\begin{figure}[!htb] 
\centering
\includegraphics[width=1.0\linewidth]{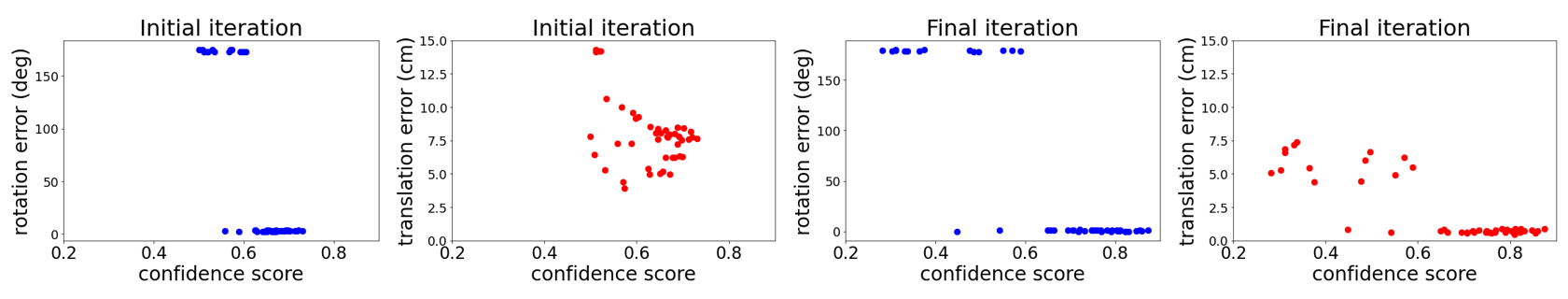}
\caption{
The correlation between confidence scores and actual camera pose errors. 
As the training progresses, images with lower pose errors tend to have higher confidence scores, leading to a major improvement of the inlier poses.
}
\label{fig:conf} 
\end{figure}

\paragraph{Why our method is robust to outliers.}
Here we provide an examination of how our method can help mitigate the influence of outliers. 
In Figure~\ref{fig:conf}, we show the confidence score distribution in the scene {\it Bear} and the pose error.
The left two subplots display the initial distributions, while the right two depict the distributions after the optimization.
Notably, images with larger pose errors exhibit lower scores, and this gap increases as the refinement progresses, during which, the inlier poses are sampled and refined, resulting in the observed improvement. 
Nevertheless, the observation in Figure~\ref{fig:conf} is just an aspect of the working mechanism of our full pipeline, which leverages a synergy of the scene graph optimization with an IoU loss and a coarse-to-fine strategy.

% \input{tab/lambda}
% \paragraph{The choice of $\lambda$.}
% The parameter $\lambda$ adjusts the impact of PSNR on the $CS$ update. In Table~\ref{tab:lambda}, we report the quantitative results on our dataset with various $\lambda$ values. The results show that as $\lambda$ increases, the performance first improves and then decreases. 
% This suggests selecting an appropriate $\lambda$ that balances the sparse (keypoint matches) and dense (photometric residuals) information. 
% Thus, we set $\lambda=1$.

\section{Conclusion}
\label{sec:conclusion}

This paper explores the task of neural surface reconstruction from image sets containing significant outlier poses. 
We introduce a novel method that jointly optimizes the neural radiance field with a scene graph. 
The key idea is to adaptively estimate the proposed inlier-outlier confidence scores and reduce the influence of outlier poses during reconstruction. 
In addition, we propose an Intersection-over-Union (IoU) loss and a coarse-to-fine strategy to facilitate the optimization process. 
Even though our method can greatly refine the inlier poses, 
the improvement on outlier poses is moderate (whose effect is still largely alleviated with the proposed confidence scheme), which we deem a limitation. Future explorations on rectifying outlier poses with visual (re)localization would be a promising direction, according to our study in this work.

\clearpage
\noindent
\textbf{Acknowledgments.} 
We thank the anonymous reviewers for their valuable feedback. 
This work is supported by 
the Early Career Scheme of the Research Grants Council (grant \# 27207224),
the HKU-100 Award, 
and in part by NSF China (No. 62172363).
Siyan Dong would also like to thank the support from HKU Musketeers Foundation Institute of Data Science for the Postdoctoral Research Fellowship.

\section{Detailed Introduction of Our Dataset}
\label{sec:supp_ourdata}

\begin{figure*}[htbp] 
\centering
\includegraphics[width=\linewidth]{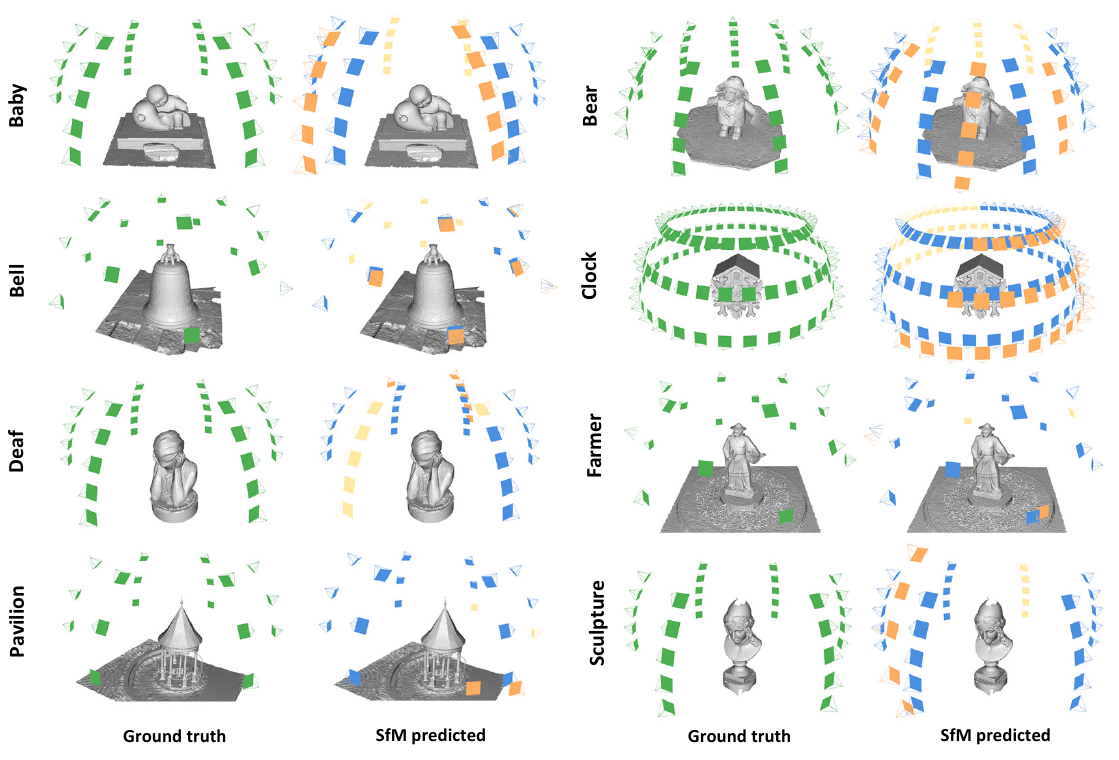}
\caption{
Illustration of the training viewpoints in our dataset. 
For each scene, the left image visualizes the ground truth camera poses (green frustums). The right image displays the SfM predicted results, which are utilized as input poses for NeRF training. The blue frustums represent poses within 20 cm and 20 deg. The orange frustums represent those larger than 20 cm and 20 deg, while the yellow frustums denote the ground truth poses of the orange ones. 
}
\label{fig:sfm} 
\end{figure*}

We investigate the problem of performing NeRF-based 3D reconstruction from images with significant pose errors. 
To benchmark the aforementioned problem, 
we collect 3D meshes from BlendedMVS~\cite{yao2020blendedmvs} and generate a new inward-facing dataset.
We select 8 representative scenes and uniformly sample camera viewpoints by different strategies in the hemisphere around each mesh.
The statistics and visualization of each scene are shown in Table~\ref{tab:data_info} and Figure~\ref{fig:sfm}. 

\begin{table}[tbh!]
\caption{
Scene statistics of our proposed dataset. 
For each scene, we report the number of total images, the rotation \& translation errors of the camera pose estimates from our SfM module, and the number of accurate poses under different tolerance thresholds. 
}
\centering
\resizebox{0.70\textwidth}{!}{

\begin{tabular}{l|c|cc|cc} 
\toprule

 & & \multicolumn{2}{|c}{Camera pose error (cm, deg)}
& \multicolumn{2}{|c}{\# $<$ thresh}  \\ 

Scene & \# Images & Mean & Median & 1, 1 & 20, 20 \\

\midrule

Baby & 30 & 3.94, 59.93 & 0.28, 0.14 & 20 & 20 \\

Bear & 45 & 3.64, 55.24 & 1.11, 0.56 & 17 & 31   \\

Bell & 18 & 10.15, 59.91 & 0.39, 0.31 & 12 & 12    \\

Clock & 108 & 14.17, 52.14 & 0.80, 0.24 & 73 & 76   \\

Deaf & 30 & 9.58, 35.95 & 1.82, 0.99 & 1 & 24  \\

Farmer & 18 & 0.75, 19.88 & 0.43, 0.33 & 15 & 16  \\

Pavilion & 18 & 2.46, 12.07 & 0.35, 0.26 & 16 & 16  \\

Sculpture & 30 & 8.29, 30.76 & 6.35, 2.62 & 0 & 25  \\

\bottomrule
\end{tabular}

}

\label{tab:data_info}
\end{table}

\begin{figure}[htbp] 
\centering
\includegraphics[width=0.62\linewidth]{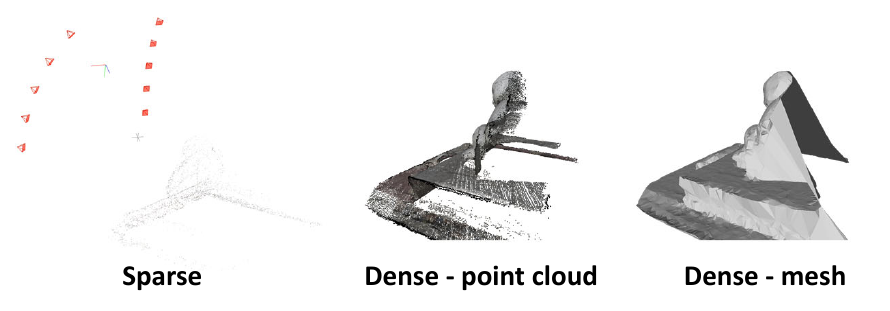}
\caption{
Illustration of COLMAP~\cite{schoenberger2016sfm} results on scene \textit{Baby}. 
Both sparse and dense reconstruction processes are unsuccessful. Only 1/3 of the viewpoints are estimated with reasonably accurate poses, and about half of the mesh is reconstructed.
}
\label{fig:colmap} 
\end{figure}

Most of the selected scenes contain 18-45 training images, with the exception of the scene \textit{Clock}, which comprises 108 training images.
We calculate the initial camera poses for the training images with our SfM module (COLMAP~\cite{schoenberger2016sfm} with SuperPoint~\cite{detone2018superpoint} and SuperGlue~\cite{sarlin2020superglue}). 
The reconstruction result of each scene contains significant incorrect poses, evident from the gap between the mean and median pose errors.
We also report the number of correct poses in each scene, with two sets of pose error thresholds: 1 cm 1 deg, and 20 cm 20 deg.

The incorrect camera pose estimates are primarily due to incorrect keypoint matches. To address this, we have enhanced COLMAP with SuperPoint and SuperGlue to create our SfM module. Through our experiments, we have consistently observed our SfM module outperforming the standard COLMAP. All experiments mentioned in the main paper, conducted on our dataset, utilized the same initial poses generated by our SfM module. 
An example of a typical COLMAP result is shown in Figure~\ref{fig:colmap}. We also showcase the dense reconstruction results with COLMAP MVS~\cite{schoenberger2016mvs}.

\paragraph{Evaluation metrics. }
Following the evaluation protocol of previous research~\cite{wang2021neus, li2023neuralangelo}, we choose Chamfer distance and F-score as metrics to evaluate 3D reconstruction quality on our dataset.
As the optimization of camera poses leads to changes in the coordinate system of the scene, it's crucial to align the obtained mesh with ground truth before evaluation. 
This is achieved by aligning the estimated camera poses with the ground truth.
Since our dataset includes outliers that could disturb the alignment process, we manually filter out camera poses with an initial error exceeding 20 cm and 20 degrees during the process. 
We first align the global orientation on top of the estimated and ground truth camera rotations. 
Then, we solve a convex optimization problem to find the global scaling factor and translation vector. 
As a result, we obtain the 7-degree-of-freedom relative transformation matrix between the estimated and ground truth camera poses. 
After the alignment, 
we scale the reconstructed and ground truth meshes by a factor of 10, sample $K=100,000$ points on each mesh surface, and calculate the metrics on top of the sampled points.

To calculate the Chamfer distance, we first compute two measurements: accuracy $Acc(\cdot)$ and completeness $Com(\cdot)$.
They are computed by: 
\begin{equation}
\begin{split}
Acc(S_{rec}, S_{gt}) = \frac{1}{K} \sum_{p \in S_{rec}} \min_{q \in S_{gt}} || p - q ||_1  ,  \\
Com(S_{rec}, S_{gt}) = \frac{1}{K} \sum_{q \in S_{gt}} \min_{p \in S_{rec}} || p - q ||_1  .
\end{split}
\end{equation}
Then, the Chamfer distance is calculated as the mean of the two aforementioned measurements: 
\begin{equation}
CD(S_{rec}, S_{gt}) = \frac{ Acc(S_{rec}, S_{gt}) + Com(S_{rec}, S_{gt}) }{2}  .
\end{equation} 

To calculate the F-score, we compute two measurements: precision $Pre(\cdot)$ and recall $Rec(\cdot)$. 
They are computed by: 
\begin{equation}
\begin{split}
Pre(S_{rec}, S_{gt}) = \frac{1}{K} \sum_{p \in S_{rec}} \max_{q \in S_{gt}} ||p - q||_0  ,  \\
Rec(S_{rec}, S_{gt}) = \frac{1}{K} \sum_{q \in S_{gt}} \max_{p \in S_{rec}} || p - q ||_0  , 
\end{split}
\end{equation}
where $||\cdot||_0$ indicates inlier/outlier points with a distance threshold $d$:
\begin{equation}
|| \cdot ||_0 = 
\left\{\begin{matrix}
1 & || \cdot ||_1 < d\\ 
0 & otherwise 
\end{matrix}\right.
\end{equation}
We set $d=0.64$. The F-score is calculated as the harmonic mean of the precision and recall: 
\begin{equation}
\begin{split}
F-score(S_{rec}, S_{gt}) & = \\ 
& \frac{ 2 \cdot Pre(S_{rec}, S_{gt}) \cdot Rec(S_{rec}, S_{gt}) }{ Pre(S_{rec}, S_{gt}) + Rec(S_{rec}, S_{gt}) }  .
\end{split}
\end{equation}

\section{More Evaluations}
\label{sec:supp_analysis}

\paragraph{Evaluation of camera pose errors.}
In Table~\ref{tab:pose}, we report mean pose errors on our dataset. 
To ensure a fair comparison, we use the confidence-weighted mean error, denoted as SG-W. This is because not all poses have an equal impact on our method. 
SG-W achieves substantial error reduction compared to competitors. 
We also report the results with hard outlier rejection by considering only the selected training viewpoints in the final epoch, denoted by SG-H. The decrease in pose errors is more significant, and 
it achieves mean precision of 68\% and recall of 80\% for outlier rejection. Here, the ground truth outliers are defined by errors $>$ 20 cm 20 deg of their initial camera poses.

\begin{table}[htb!]

\caption{
Evaluation of camera pose errors on the proposed dataset. 
}

\centering
\resizebox{0.96\textwidth}{!}{ 
\begin{tabular}{l|ccccccc|ccc}

\toprule 

Mean errors & BARF & SCNeRF & GARF & L2G-NeRF & Joint-TensoRF & CamP & PoRF & SG & SG-W & SG-H \\ 

\toprule 

Translation (cm) & 0.71 & 0.69 & 0.78 & 0.82 & 0.88 & 0.69 & 0.75 & 0.70 & 0.45 & 0.09 \\

\midrule

Rotation (deg) & 
39.40 &
40.70 &
47.07 &
45.31 &
46.55 & 
40.82 &
43.26 & 
39.95 &
25.73 &
4.62 \\

\bottomrule

\end{tabular}

}

\label{tab:pose}
\end{table}

\paragraph{Comparison with recent camera pose optimizer CamP~\cite{park2023camp}. }
We have tested CamP in the scene {\it Bell}. The mean camera pose error is 1.01 cm 60.04 deg. 
We find that CamP does not always guarantee the prevention of the optimization falling into local minima, especially when there are outlier camera poses. 
As a result, the pose errors are similar to those of SCNeRF.

\begin{figure*}[!ht] 
\centering
\includegraphics[width=1.0\linewidth]{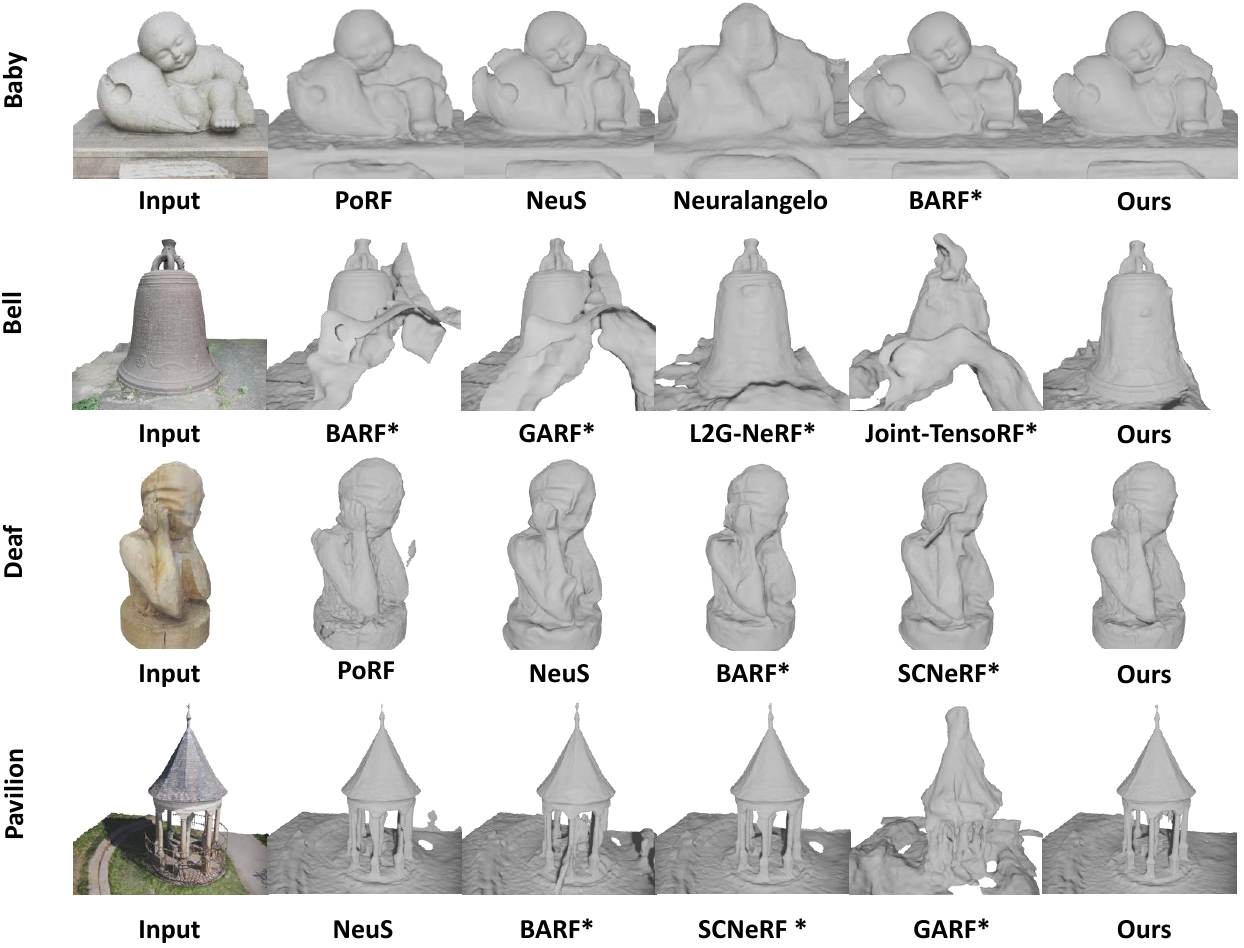}
\caption{
Additional qualitative comparison on our dataset. 
}
\label{fig:ours_supp} 
\end{figure*}

\paragraph{Additional qualitative comparison on our dataset. }
Due to the page number limit, we show qualitative results for only 4 scenes in Figure~\ref{fig:teaser} and Figure~\ref{fig:ours} in the main paper. The remaining results can be found in Figure~\ref{fig:ours_supp}.

\paragraph{Discussion and comparison with SPARF~\cite{truong2023sparf}. }
\textbf{Problem setting}: 
SPARF tackles the problem of few-shot reconstruction (typically 3 views) with moderate camera poses; we focus on the common inward-facing scenario that can include real large pose errors. 
\textbf{Method design}: 
While both SPARF and our method utilize multi-view keypoint correspondences, we introduce an IoU formulation. This allows us to enhance the implicit geometry in a more continuous optimization space (i.e., aligning MoGs), as opposed to directly applying the conventional re-projection error.
Additionally, we do not require dense depth rendering, which can be computationally expensive and consume a significant amount of memory. 
\textbf{Quantitative comparison}: 
We have tested SPARF on the scene {\it Bell}. 
Through the experiment, we find that the outliers' matches and poses can significantly mislead SPARF's optimization process. This results in a mean pose error of 1.17 cm 165.19 deg (SG-W achieves 0.85 cm and 43.09 deg).

\paragraph{Upper bound. }
To better understand the reconstruction results on our dataset, we provide an upper bound estimation on top of our method.
We utilize the ground truth camera poses and set these poses as fixed.
The results are shown in Table~\ref{tab:upper}. 
Our method has demonstrated significant improvements compared to plain NeuS~\cite{wang2021neus}. However, there is still room for further research to enhance performance. Although our approach, which involves the soft rejection of outliers, mitigates the influence of these outliers, it doesn't completely eliminate them. These outliers can still affect the reconstruction performance. 
We believe that combining our method with hard outlier rejection or visual (re)localization techniques could potentially address this issue. This presents a promising direction for future research.

\begin{table*}[htbp]
\caption{
Additional quantitative results on our dataset. 
}

\centering
\resizebox{0.96\textwidth}{!}{ 

\begin{tabular}{cl|cccccccc|c}

\toprule
 & & Baby & Bear & Bell & Clock & Deaf & Farmer & Pavilion & Sculpture & Mean \\ 

\midrule
 
\multirow{9}{*}{\rotatebox{90}{ \scriptsize{Chamfer distance $\downarrow$}} } 

& CasMVSNet~\cite{gu2020cascade}  & 1.04 & 0.78 & 1.93 & 1.24 & 1.04 & 3.25 & 1.16 & 1.12 & 1.45 \\ 

& IterMVS~\cite{wang2022itermvs} & 0.73 & 0.65 & 1.51 & 0.90 & 0.86 & 1.70 & 0.90 & 0.57 & 0.98 \\ 

\cline{2-11}

& BARF~\cite{lin2021barf}\# & 0.72 & 0.40 & 5.11 & 0.26 & 0.49 & 5.73 & 5.29 & 0.68 & 2.34 \\ 

& SCNeRF~\cite{jeong2021self}\# & 0.71 & 0.64 & - & 0.49 & 0.50 & 5.18 & 2.60 & 0.58 & 1.53$^\dagger$ \\ 

& GARF~\cite{chng2022gaussian}\# & - & 0.47 & - & 2.21 & 0.89 & 5.37 & - & 0.59 & 1.91$^\dagger$ \\ 

& L2G-NeRF~\cite{chen2023local}\# & 0.74 & 1.85 & 6.82 & 0.35 & 0.53 & 6.04 & 3.12 & 2.45 & 2.74 \\ 

& Joint-TensoRF~\cite{cheng2024improving}\# & - & 1.34 & - & 0.73 & 1.30 & 6.23 & 3.35 & 1.45 & 2.40$^\dagger$ \\

\cline{2-11}

& SG-NeRF (Ours) & 0.56 & 0.25 & 0.98 & 0.15 & 0.45 & 0.87 & 0.20 & 0.22 & 0.46 \\ 

& Upper bound & 0.37 & 0.08 & 0.23 & 0.08 & 0.31 & 0.19 & 0.14 & 0.16 & 0.20 \\ 

\midrule

\multirow{9}{*}{\rotatebox{90}{\scriptsize{F-score $\uparrow$}}} 

& CasMVSNet~\cite{gu2020cascade} & 0.43 & 0.57 & 0.33 & 0.39 & 0.40 & 0.27 & 0.51 & 0.46 & 0.42 \\ 

& IterMVS~\cite{wang2022itermvs} & 0.53 & 0.65 & 0.39 & 0.53 & 0.41 & 0.38 & 0.57 & 0.62 & 0.51 \\ 

\cline{2-11}

& BARF~\cite{lin2021barf}\# & 0.67 & 0.87 & 0.08 & 0.90 & 0.76 & 0.10 & 0.41 & 0.81 & 0.58 \\ 

& SCNeRF~\cite{jeong2021self}\# & 0.54 & 0.79 & - & 0.78 & 0.77 & 0.12 & 0.37 & 0.69 & 0.58$^\dagger$ \\ 

& GARF~\cite{chng2022gaussian}\# & - & 0.83 & - & 0.11 & 0.46 & 0.12 & - & 0.80 & 0.46$^\dagger$ \\ 

& L2G-NeRF~\cite{chen2023local}\# & 0.68 & 0.65 & 0.02 & 0.86 & 0.79 & 0.10 & 0.22 & 0.16 & 0.44 \\ 

& Joint-TensoRF~\cite{cheng2024improving}\# & - & 0.35 & - & 0.57 & 0.22 & 0.08 & 0.15 & 0.25 & 0.27$^\dagger$ \\ 

\cline{2-11}

& SG-NeRF (Ours) & 0.74 & 0.93 & 0.71 & 0.96 & 0.87 & 0.76 & 0.94 & 0.92 & 0.85 \\ 

& Upper bound & 0.80 & 0.99 & 0.92 & 0.99 & 0.91 & 0.95 & 0.97 & 0.94 & 0.93 \\

\bottomrule

\end{tabular}

}

\label{tab:upper}
\end{table*}

\paragraph{Neuralangelo with pose optimization.}
In the main paper, we utilize NeuS~\cite{wang2021neus} as the backbone for 3D reconstruction and compare various pose optimization methods. In table~\ref{tab:upper}, we present the results of combining the pose optimization methods with Neuralangelo~\cite{li2023neuralangelo}. The methods are identified as BARF~\cite{lin2021barf}\#, SCNeRF~\cite{jeong2021self}\#, and so on. 
As observed, the results are worse than those with NeuS, since Neuralangelo is more sensitive to outliers. 
The results further confirm our effectiveness in handling outliers. They also demonstrate the non-trivial design for joint optimization of the radiance field and camera poses, instead of simply combining different methods.

\paragraph{Comparison with MVS methods.}
Since our focus is on neural surface reconstruction with radiance fields, we do not compare our work with classical Multi-View Stereo (MVS) algorithms in the main paper.
Here, in table~\ref{tab:upper}, we report the results from two recent MVS methods: CasMVSNet~\cite{gu2020cascade} and IterMVS~\cite{wang2022itermvs}. 
As MVS-based methods generally utilize the epipolar geometry prior that heavily relies on camera poses, these methods are sensitive to pose errors.

\paragraph{Additional comparison with more surface reconstruction methods. } 
Since our goal is to mitigate the impact of outliers, we primarily compare with pose optimization methods in the main paper.
Here, we also run HF-NeuS~\cite{wang2022hf}, Voxsurf~\cite{wu2022voxurf}, and NeuDA~\cite{cai2023neuda} on the scene {\it Bell}. The Chamfer distances (1.41, 2.64, and 1.72) are larger than our method. 
We notice that these methods that aim to reconstruct details tend to be sensitive to large pose errors.

\paragraph{Bundle adjustment (BA) with the pruned scene graph.} 
We have tried to run BA on the pruned graph, but the pose updates are minimal. Note that the removed edges by our pruning step were considered good and consistent during previous COLMAP processing (including BA). 
Since BA does not use residuals on dense pixel colors, the new round of BA on the sparse graph makes only a small contribution. As a result, NeuS still produces unsatisfactory results.

\begin{wraptable}{r}{0.5\textwidth}

\caption{
Results with different $\lambda$. 
}

\centering
\resizebox{0.50\textwidth}{!}{

\begin{tabular}{l|cccccc} 
\toprule

$\lambda$
& 0.2 & 0.5 & 1.0 & 1.5 & 2.0 & 3.0 \\ 
\midrule

Camfer $\downarrow$ 
& 0.5058 & 0.4820 & \textbf{0.4646} & 0.4788 & 0.4885 & 0.4818 \\

F-score $\uparrow$ 
& 0.8379 & 0.8516 & \textbf{0.8524} & 0.8461 & 0.8438 & 0.8384 \\

\bottomrule
\end{tabular}
}

\label{tab:lambda}
\end{wraptable}
\paragraph{The choice of $\lambda$.}
The parameter $\lambda$ adjusts the impact of PSNR on the $CS$ update. In Table~\ref{tab:lambda}, we report the quantitative results on our dataset with various $\lambda$ values. The results show that as $\lambda$ increases, the performance first improves and then decreases. 
This suggests selecting an appropriate $\lambda$ that balances the sparse (keypoint matches) and dense (photometric residuals) information. 
Thus, we set $\lambda=1$.

\section{Real-World Scene Reconstruction}
\label{sec:supp_real}

\begin{wrapfigure}{r}{0.5\textwidth}
\centering
\includegraphics[width=0.98\linewidth]{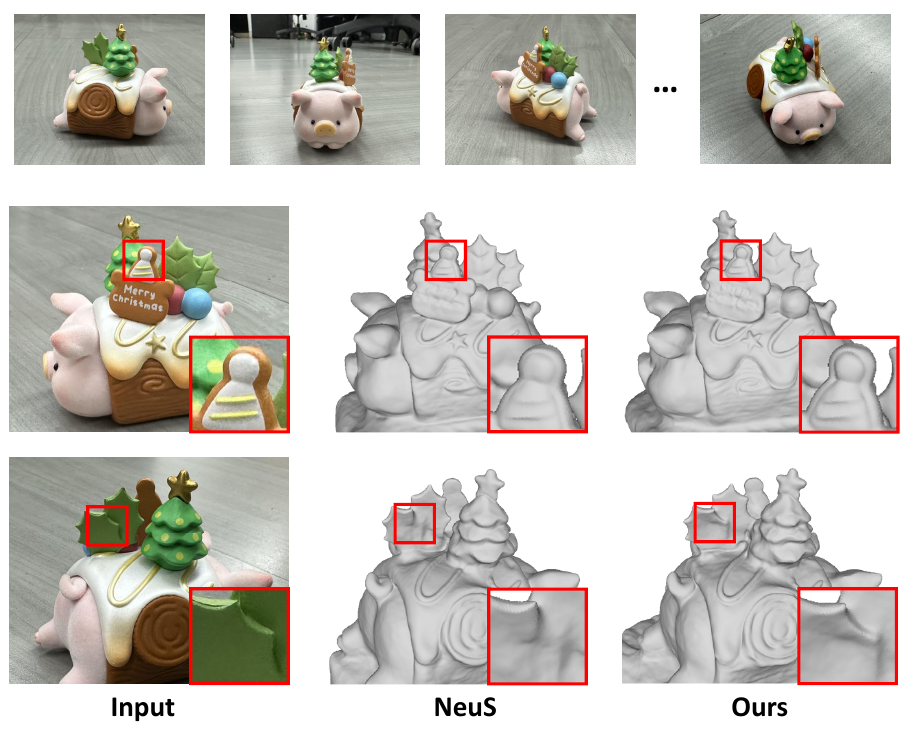}
\caption{
Qualitative comparison of a real-world scene reconstruction.
}
\label{fig:real} 
\end{wrapfigure}

In this paper, we tackle a challenging, yet practical scenario where images are casually captured without careful selection. To highlight the benefits of our method under such conditions, we showcase a real-world reconstruction example.
We casually capture 20 images of a toy object in an inward-facing manner. While COLMAP fails to solve the camera poses for this scene, our SfM module estimates the poses reasonably. 
Given the initial camera poses, we train NeuS~\cite{wang2021neus} and our method and then extract meshes for a quantitative comparison. 
Figure~\ref{fig:real} illustrates the results. 
We can observe that NeuS's mesh contains wrong geometries and tends to be over-smoothed. 
On the contrary, our reconstruction presents correct structures and more details, demonstrating our robustness to wrong poses.

\bibliographystyle{splncs04}
%\bibliography{main}

\end{document}